\documentclass{article}
\usepackage{arxiv_preprint,times}
\usepackage{hyperref}
\usepackage{url}
\usepackage{graphicx}
\usepackage{amsmath}
\usepackage{algorithm}
\usepackage{algorithmic}
\usepackage{booktabs}
\usepackage{colortbl}
\usepackage{multirow}
\usepackage{float}

\usepackage{wrapfig}
\usepackage{graphicx}

\usepackage{booktabs}
\usepackage{amssymb}
\usepackage{tabularx}
\usepackage{array}
\newcolumntype{C}[1]{>{\centering\arraybackslash}p{#1}}
\usepackage{makecell}
\usepackage{caption}

\title{ComManip: Overfitting Manipulation Policies to Comfortable Regions}
\author{Lai Yidan\textsuperscript{1}\thanks{Email: \texttt{202364820522@mail.scut.edu.cn}},
Xinyi Chen\textsuperscript{1}\thanks{Email: \texttt{202511096274@mail.scut.edu.cn}},
Shuquan Man\textsuperscript{2}\thanks{Email: \texttt{wimanshuquan@mail.scut.edu.cn}}, and
Huiping Zhuang\textsuperscript{1}\thanks{Corresponding author: \texttt{hpzhuang@scut.edu.cn}}\\
\textsuperscript{1}South China University of Technology, Guangzhou, China\\
\textsuperscript{2}Nanyang Technological University, Singapore}

\begin{document}

\maketitle

\begin{abstract}
Training robot manipulation policies relies on costly robot demonstrations, making large-scale data collection impractical. Meanwhile, to improve policy generalization, existing approaches seek greater diversity in visual observations by varying object placements, viewpoints, and robot configurations during data collection. However, under a limited demonstration budget, this strategy forces the policy to model diverse visual observations, providing insufficient supervision to learn reliable observation–action correspondences under similar local conditions. Our study reveals that policies trained under this strategy achieve lower task success rates than those trained within a compact, visually and kinematically stable region. We refer to these stable regions as comfortable manipulation regions. To exploit this finding, we propose ComManip, a learning paradigm that specializes manipulation policies to comfortable manipulation regions. During inference, ComManip repositions the mobile base until the detected target center enters the familiar image-space range estimated from comfortable-region demonstrations. It then executes the same manipulation policy, enabling effective manipulation across diverse target locations. We conduct extensive experiments across multiple manipulation tasks, demonstration budgets, and policy families including ACT, $\pi$0.5, RDT, OpenVLA-OFT, and SmolVLA. The results demonstrate that ComManip improves task success by roughly 20 percentage points or more across different policy architectures in large workspaces under limited demonstration budgets, suggesting that specializing manipulation policies to comfortable regions provides a more data-efficient learning paradigm for manipulation.
\end{abstract}

\section{Introduction}
Robot manipulation is rapidly evolving from task-specific systems toward general policies capable of performing diverse manipulation skills across a wide range of tasks and environments~\citep{octo2024,kim2025openvla,black2025pi05}. Training such general-purpose policies requires large and diverse robot datasets that cover varied operating conditions. However, collecting robot demonstrations remains costly and time-consuming, as each demonstration requires task execution, environment reset, and quality control~\citep{pmlr-v229-walke23a,khazatsky2024droid}. Improving the data efficiency of policy learning under limited demonstration budgets is therefore critical.

Beyond data efficiency, manipulation policies must generalize across varying deployment conditions. To improve generalization, existing approaches vary object placements, viewpoints, and robot configurations during data collection to cover the visual and kinematic variations expected at deployment. Among these factors, tabletop manipulation commonly emphasizes spatial diversity in target placement, aiming to enable policies to operate across diverse target locations during deployment~\citep{gao2024efficient,chen2026robotwin}.

However, the effectiveness of target-placement diversity depends critically on the available demonstration budget. Under a limited budget, sampling more target locations reduces the demonstration density within each local region. The same number of demonstrations must therefore capture a broader range of visual observations and manipulation trajectories, providing fewer examples under similar local conditions. This raises the question of whether such diversity remains beneficial when demonstrations are limited.

\begin{figure}[t]
    \centering
    \includegraphics[width=0.95\linewidth]{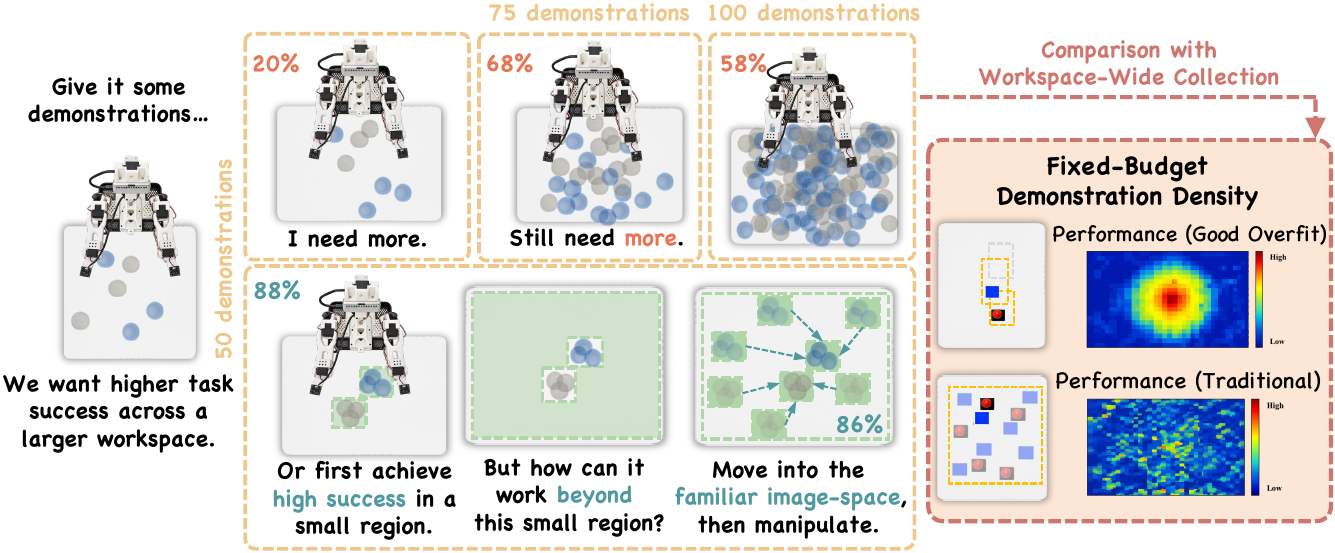}
    \caption{Empirical observation motivating this work. Under the same demonstration budget, concentrating demonstrations within a compact local region yields higher task success rates than distributing demonstrations to cover a larger set of possible object locations.}
    \label{figphenomenon}
\end{figure}

To better understand this issue, we examine how the spatial distribution of demonstrations affects policy learning under a fixed demonstration budget. As shown in Figure~\ref{figphenomenon}, under the same budget, concentrating demonstrations within a compact local region consistently yields higher task success rates than distributing demonstrations to cover more object locations. This observation suggests that, when demonstrations are limited, manipulation policies are more effectively learned as locally specialized skills than as a single policy expected to directly handle broad spatial variation. We refer to these compact regions in the physical workspace where limited demonstrations support reliable policy learning as \emph{comfortable manipulation regions}.

Motivated by this observation, we propose ComManip, a learning paradigm that intentionally trains locally specialized manipulation policies within comfortable regions rather than expanding the training distribution across the entire workspace. At deployment, ComManip repositions the mobile base until an out-of-region target center enters the familiar image-space range associated with the training demonstrations. It then executes the same manipulation policy, enabling effective manipulation across diverse target locations.

Our contributions are summarized as follows:
\begin{itemize}
    \item We show that directly learning from spatially diverse target placements is less data-efficient than learning a locally specialized policy within a comfortable manipulation region under limited demonstration budgets.
    \item We propose ComManip, which combines comfortable-region specialization with visual alignment to the familiar image-space range through repositioning, enabling manipulation across diverse target locations.
    \item We conduct extensive experiments across manipulation tasks, demonstration budgets, and policy families, including ACT, $\pi$0.5, RDT, OpenVLA-OFT, and SmolVLA, demonstrating that ComManip consistently improves performance under diverse target placements.

\end{itemize}

\section{Related Work}
\paragraph{Data-Driven Robot Manipulation.}
Large-scale robot datasets have enabled policy learning across increasingly diverse tasks, environments, and robot embodiments~\citep{oneill2024openxembodiment,pmlr-v229-walke23a,khazatsky2024droid}. Generalist policies and vision-language-action (VLA) models further leverage such data through language conditioning, large-scale pretraining, and expressive action modeling~\citep{octo2024,brohan2023rt2,kim2025openvla,black2025pi05,liu2025rdt,shukor2025smolvla}. At the policy-modeling level, ACT and Diffusion Policy improve action prediction through action chunking and generative trajectory modeling, respectively~\citep{zhao2023learning,chi2023diffusion}. These studies have advanced manipulation policy learning through larger and more diverse datasets as well as improved policy architectures.

\paragraph{Data Diversity and Local Policy Learning.}
Manipulation datasets commonly vary object placements, camera viewpoints, robot configurations, and scene layouts to improve policy generalization across operating conditions~\citep{gao2024efficient,chen2026robotwin}. However, under a limited demonstration budget, covering more operating conditions spreads demonstrations more thinly, reducing the number of examples available for each condition. ACA~\citep{chen2026aca} addresses this tradeoff by collecting repeated demonstrations around selected anchor configurations and selectively expanding toward boundaries with high risk. ReconVLA~\citep{song2026reconvla} reconstructs gaze regions of target objects to enhance perception relevant to manipulation, while PALM~\citep{wang2026palm} exploits invariant local action distributions and consistent visual and proprioceptive representations to reduce discrepancies across distributions. These methods improve generalization through data collection or representation learning. In contrast, ComManip concentrates demonstrations within a compact region and addresses workspace variation through base repositioning during deployment, without modifying the policy architecture or learning objective.

\paragraph{Mobile Base Repositioning for Manipulation.}
The mobile base pose influences both visual observations and manipulation feasibility. Existing methods improve manipulation through base pose selection or coordinated base and arm motion, using scene reconstruction and pose optimization~\citep{yang2025mobipi}, top-down spatial representations~\citep{naik2025vbmnet}, navigation or semantic affordances~\citep{zhang2025momakitchen,lin2026affordance}, reachability-aware trajectory optimization~\citep{wu2025momanipvla}, or pose preferences learned from manipulation rollouts~\citep{chai2026n2m}. Mobi-$\pi$~\citep{yang2025mobipi}, in particular, repositions the robot to recover observations compatible with the training distribution of a pretrained manipulation policy. These studies primarily address how to select suitable base poses for a given policy, whereas our work investigates the role of deployment repositioning in improving policy learning under limited demonstration budgets.
\begin{figure*}[t]
    \centering
    \includegraphics[width=1.00\textwidth]{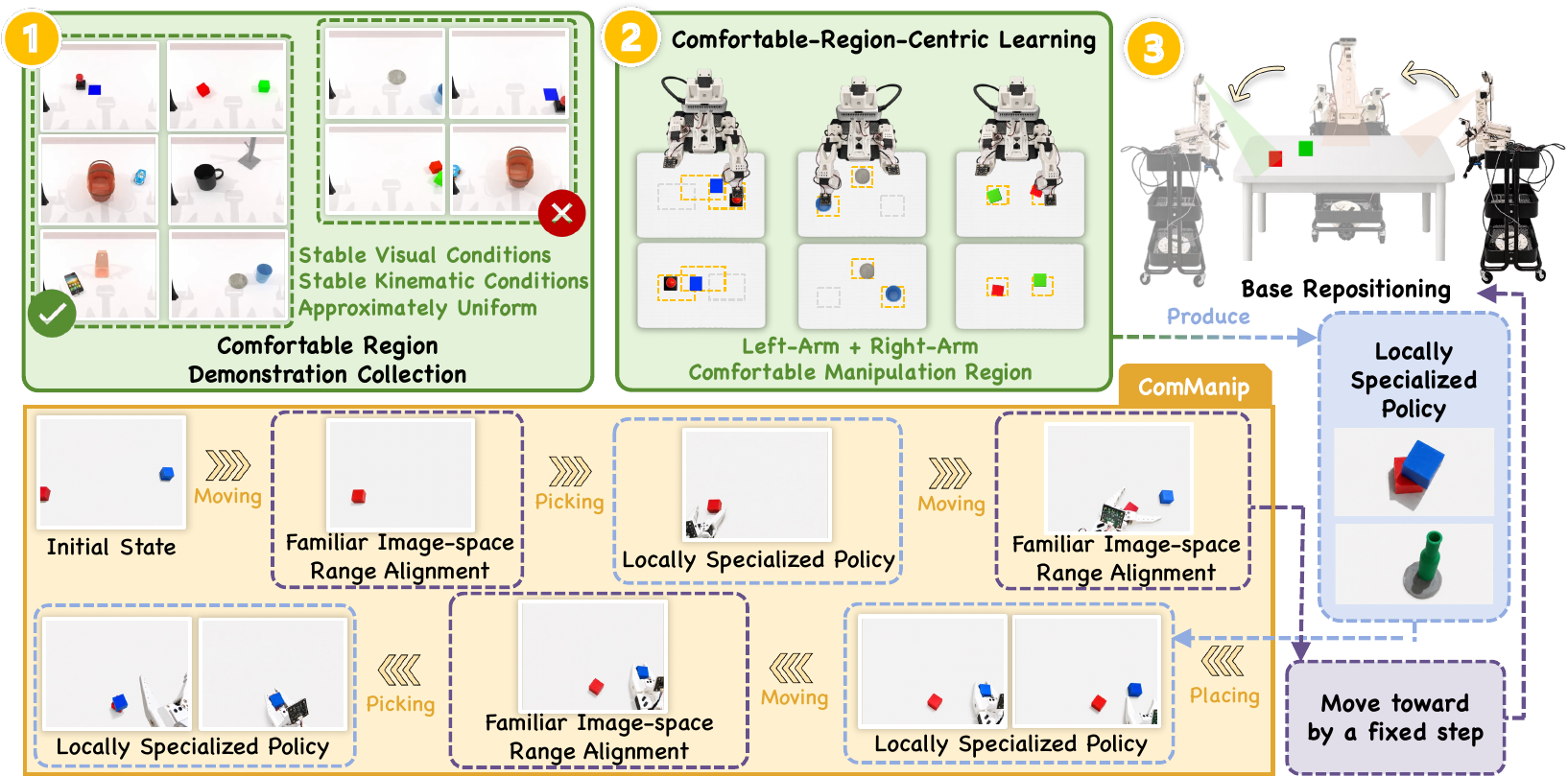}
   \vspace{-4mm}
    \caption{\textbf{Overview of ComManip.}
    (1) Demonstrations are collected within a comfortable manipulation region under stable visual and kinematic conditions.
    (2) A locally specialized policy is trained on these demonstrations.
    (3) During deployment, the mobile base is repositioned to align the target with the familiar image-space range before policy execution.}
    \label{figoverview}
    
\end{figure*}
\section{Method}

\subsection{Framework Overview}
ComManip enables a manipulation policy trained on comfortable-region demonstrations to retain its local performance when target objects are placed outside the comfortable manipulation region across the tabletop workspace. During deployment, ComManip repositions the mobile robot until the detected target center enters the familiar image-space range estimated from comfortable-region demonstrations. The same locally specialized policy is then executed. Figure~\ref{figoverview} summarizes the framework, consisting of comfortable-region demonstration collection, local policy learning, and deployment-time base repositioning. Together, these components extend a locally trained policy to diverse target placements across the workspace.

\subsection{Comfortable Manipulation Region}
\label{sec:comfortable-region}

We define a comfortable manipulation region $\mathcal{R}_C$ as a
compact region in the physical workspace in which the target is
clearly visible and the robot operates under favorable kinematic
conditions, thereby providing stable conditions for local policy
learning. For each task, we select a reference target location that
is clearly visible and reachable by the robot under the given setup, and specify a task-dependent spatial extent around it. As shown in the comfortable-region collection panels of
Figure~\ref{figoverview}, the resulting region covers spatially varied target placements while maintaining favorable visual and kinematic conditions. Target placements are approximately uniformly sampled within this region, from which we retain $N$ successful demonstrations.
The same demonstrations are used for policy training and for estimating
the target-center statistics required during deployment, so no separate
calibration set is needed.

To characterize the image-space target locations observed in demonstrations collected within $\mathcal{R}_C$, GroundingDINO~\citep{liu2024groundingdino} and SAM2~\citep{ravi2024sam2} are applied to the initial observation $o_{i,0}$ of each retained demonstration. GroundingDINO localizes the target object, and SAM2 produces its segmentation mask. We extract the image-space center of the target as
\begin{equation}
    c_i
    =
    \Phi(o_{i,0})
    =
    (x_i,y_i),
\end{equation}
where $\Phi$ denotes the target localization, segmentation, and center-extraction process. The resulting image-space target centers form the training
target-center set
\begin{equation}
\mathcal{C}_C = \{c_i\}_{i=1}^{N}.
\end{equation}

Before estimating $\mathcal{F}_C$ and $\mu_C$, we manually
remove a few isolated centers far from the main concentration;
the training demonstrations remain unchanged. The retained target-center set is
\begin{equation}
\widetilde{\mathcal{C}}_C
\subseteq
\mathcal{C}_C.
\end{equation}

We compute the empirical mean of the retained target centers as
\begin{equation}
\mu_C =
\frac{1}{\left|\widetilde{\mathcal{C}}_C\right|}
\sum_{c_i\in\widetilde{\mathcal{C}}_C} c_i,
\end{equation}
where $\mu_C$ is the empirical mean of the retained training
target centers in image space.

The physical comfortable manipulation region $\mathcal{R}_C$ specifies where initial target placements are sampled. Under the fixed initial robot and camera configuration used for data
collection, these placements yield the image-space target centers that form $\mathcal{C}_C$. After lightweight trimming, the retained set $\widetilde{\mathcal{C}}_C$ is used to compute the empirical mean $\mu_C$ and estimate the familiar range
$\mathcal{F}_C$. Therefore, $\mathcal{F}_C$ serves as an empirical image-space proxy for the concentration of target locations observed in demonstrations collected within
$\mathcal{R}_C$, rather than a geometrically equivalent representation of $\mathcal{R}_C$. These quantities are used in Sec.~\ref{sec:workspace-expansion} for deployment-time visual alignment.

\subsection{Comfortable-Region-Centric Learning}
The $N$ successful demonstrations collected within the selected
comfortable manipulation region constitute the training dataset
\begin{equation}
\mathcal{D}_C=\{\xi_i\}_{i=1}^{N},
\end{equation}
where $\xi_i=\{(o_{i,t},a_{i,t})\}_{t=1}^{T_i}$ denotes the $i$-th demonstration, $T_i$ is its trajectory length, and $o_{i,t}$ and $a_{i,t}$ are the observation and corresponding robot action at time step $t$, respectively. Here, $o_{i,t}$ is used as a unified notation for the policy input. Its exact composition follows the original input specification of each policy family. Unlike conventional workspace-wide data collection, ComManip allocates the fixed budget of $N$ successful demonstrations to the selected comfortable manipulation region rather than distributing the same budget over a larger workspace.

The manipulation policy $\pi_C$ is then trained using the original learning objective of the underlying policy family,
\begin{equation}
\pi_C
=
\operatorname*{arg\,min}_{\pi}
\mathcal{L}_{\mathrm{policy}}(\pi,\mathcal{D}_C),
\end{equation}
where $\mathcal{L}_{\mathrm{policy}}$ denotes the
policy-specific training objective. For each policy--task pair,
the policy is trained on its corresponding comfortable-region
dataset $\mathcal{D}_C$ using the demonstration budget
specified in Sec.~\ref{sec:exp-setup}. Each policy family retains its original architecture, input specification, optimization objective, and training procedure. 

This strategy favors reliable local specialization over
workspace-wide invariance. Workspace variation is handled
during deployment through base repositioning, allowing
policy learning to focus on precise local manipulation.

\begin{figure}[t]
\centering
\includegraphics[width=\columnwidth]{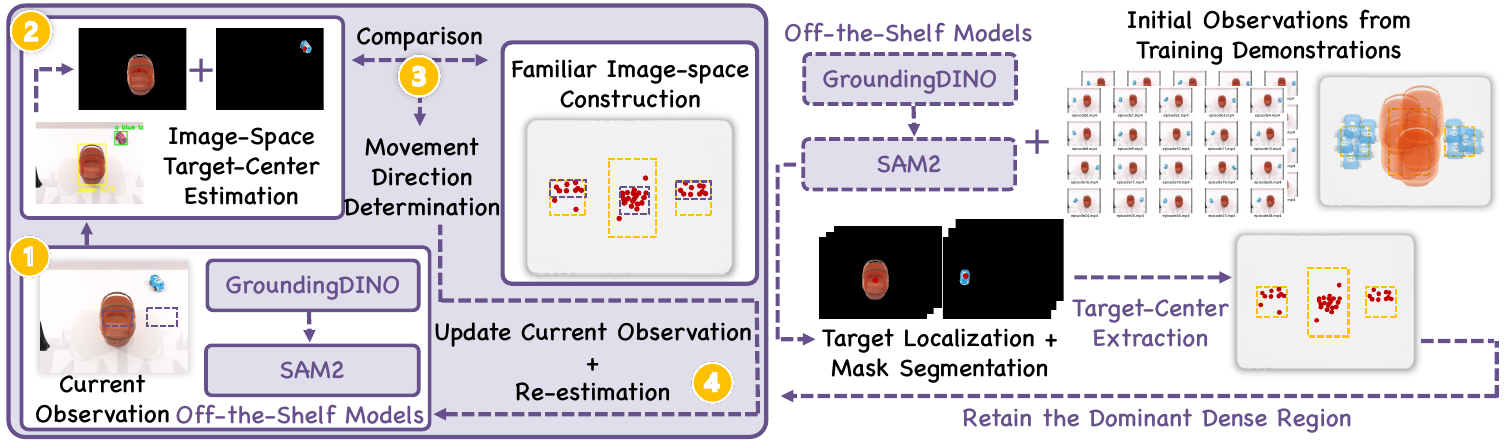}
\vspace{-4mm}
\caption{\textbf{Visual alignment for base repositioning.}
GroundingDINO and SAM2 extract target centers from the initial observations of training data, from which the retained target centers are used to estimate the familiar image-space range $\mathcal{F}_C$ and compute the empirical target-center mean $\mu_C$. During deployment, the image-space target center is estimated and compared with $\mathcal{F}_C$. The robot moves the base by a fixed task-specific step in a direction that reduces the image-space deviation from $\mu_C$ and re-estimates the target center, repeating this process until it enters $\mathcal{F}_C$.}
\label{fig:visual-alignment}
\end{figure}

\subsection{Mobile Base Repositioning}
\label{sec:workspace-expansion}

The policy $\pi_C$ is trained on demonstrations whose initial target placements are sampled within the physical comfortable manipulation region $\mathcal{R}_C$. At deployment, ComManip does not explicitly determine whether the target lies within $\mathcal{R}_C$. Instead, whether the detected image-space target center lies within $\mathcal{F}_C$ is used as the visual alignment criterion. If the detected target center lies outside $\mathcal{F}_C$, ComManip repositions the mobile robot until the center enters this familiar image-space range. Recall that $\mathcal{C}_C$ denotes the complete training target-center set extracted from the initial observations of
the training demonstrations, while
$\widetilde{\mathcal{C}}_C \subseteq \mathcal{C}_C$
denotes the retained target centers after lightweight trimming.
We define the familiar range from
$\widetilde{\mathcal{C}}_C$ as
\begin{equation}
\mathcal{F}_C =
\left\{
c=(x,y)
\;\middle|\;
x_{\min}\leq x\leq x_{\max},
\;
y_{\min}\leq y\leq y_{\max}
\right\},
\end{equation}
where
\begin{equation}
\begin{aligned}
x_{\min}
&=
\min_{c_i\in\widetilde{\mathcal{C}}_C} x_i,
&
x_{\max}
&=
\max_{c_i\in\widetilde{\mathcal{C}}_C} x_i,\\
y_{\min}
&=
\min_{c_i\in\widetilde{\mathcal{C}}_C} y_i,
&
y_{\max}
&=
\max_{c_i\in\widetilde{\mathcal{C}}_C} y_i.
\end{aligned}
\end{equation}
Thus, the resulting rectangle provides a simple empirical
image-space bound on the main concentration of target centers
observed in the training demonstrations and is used as the
familiar range $\mathcal{F}_C$ for deployment-time visual
alignment.

As shown in Figure~\ref{fig:visual-alignment}, base repositioning is performed through a closed-loop visual alignment process. At deployment, GroundingDINO detects the target specified by the task prompt, and SAM2 extracts its segmentation mask. The center of the resulting mask is denoted by
\begin{equation}
c_{\mathrm{test}}=(x_{\mathrm{test}},y_{\mathrm{test}}).
\end{equation}
For tasks involving multiple relevant objects, the same
alignment procedure is applied sequentially to each target
using its corresponding familiar range and target-center mean;
for notational simplicity, these are denoted by
$\mathcal{F}_C$ and $\mu_C$ for the target currently being
processed.

When $c_{\mathrm{test}}\notin\mathcal{F}_C$, ComManip moves the robot base by a fixed task-specific step along the table edge in the direction that reduces the image-space deviation between $c_{\mathrm{test}}$ and $\mu_C$. A new observation is then acquired and the target center is estimated again. This process continues until $c_{\mathrm{test}} \in \mathcal{F}_C$ or the maximum number of base movements $K_{\max}$ is reached. During repositioning, the robot maintains the initial arm
configuration used for policy training. Once alignment is complete, the base stops and $\pi_C$ is executed without modification.

Algorithm~\ref{algcomomanip} summarizes this procedure. The alignment rule does not estimate an optimal base pose and is compatible with different manipulation policy architectures.

\begin{algorithm}[t]
\caption{ComManip Deployment}
\label{algcomomanip}
\begin{algorithmic}[1]
\STATE Load $\pi_C$, $\mathcal{F}_C$, $\mu_C$, and $K_{\max}$
\STATE Acquire the current observation
\STATE Estimate the target center $c_{\mathrm{test}}$
\STATE $k \leftarrow 0$
\WHILE{$c_{\mathrm{test}} \notin \mathcal{F}_C$
       and $k < K_{\max}$}
    \STATE Determine the base movement direction from
    $c_{\mathrm{test}}-\mu_C$
    \STATE Move the base by one task-specific step
    \STATE $k \leftarrow k+1$
    \STATE Acquire a new observation
    \STATE Re-estimate $c_{\mathrm{test}}$
\ENDWHILE
\STATE Execute $\pi_C$
\end{algorithmic}
\end{algorithm}

\section{Experiments}
We first study the effect of target placement diversity on data efficiency under limited demonstration budgets (Sec.~\ref{sec:exp-motivate}), which motivates ComManip. The remaining experiments address four questions.
\begin{itemize}
\item Does ComManip improve performance across tasks and policies? (Sec.~\ref{sec:exp-policies})
\item Does visual alignment preserve local performance across diverse placements? (Sec.~\ref{sec:exp-alignment})
\item How does comfortable-region extent affect performance? (Sec.~\ref{sec:exp-spatial})
\item How does demonstration quantity affect performance? (Sec.~\ref{sec:exp-data})
\end{itemize}

\subsection{Experimental Setup}
\label{sec:exp-setup}

\paragraph{Simulation Evaluation.}
We evaluate ComManip in RoboTwin~2.0~\cite{chen2026robotwin} on six tasks: Stamp Seal, Empty Cup, Object Basket, Phone Stand, Hanging Mug, and Stack Blocks. The tasks cover single- and dual-arm manipulation, including stamping, grasp-and-place, constrained placement, hanging, and multi-object stacking. Base repositioning is emulated along the tabletop edge using fixed task-specific steps until the target center enters the familiar image-space range.

\paragraph{Real World Evaluation.}
We validate ComManip on Bottle Placing and Blocks Stacking using a robot mounted on an XLeRobot mobile base~\cite{wang2025xlerobot}. The base performs the same repositioning procedure along the workspace edge. In the real-world experiments, each task is evaluated over 30 trials per method.

\paragraph{Data Collection and Policies.}
Demonstrations are generated using RoboTwin's automated expert pipeline. We evaluate ACT~\cite{zhao2023learning}, $\pi_{0.5}$~\cite{black2025pi05}, RDT~\cite{liu2025rdt}, OpenVLA-OFT~\cite{kim2025openvlaoft}, and SmolVLA~\cite{shukor2025smolvla}. Each policy follows its original architecture and training procedure, with task-specific configurations. For each policy and task, the baseline and ComManip use identical training settings. See Appendix~\ref{app:training-details} for details.
\paragraph{Metrics.}
We report task success rates under the predefined success criterion for each task. For the main policy comparison in Table~\ref{tabmainresults}, we train five independent runs with different random seeds and evaluate each run on 100 episodes. We report the mean success rate across the five runs.

\subsection{Evaluation Protocol and Compared Settings}
\label{sec:exp-protocol}

We distinguish the familiar image-space range $\mathcal{F}_C$ from the physical target-placement ranges used for evaluation. We denote the comfortable region $\mathcal{R}_C$ as $\mathrm{C}$ and the broader workspace range extending toward the tabletop boundaries as $\mathrm{W}$. The notation $\mathrm{X}\!\rightarrow\!\mathrm{Y}$ indicates training in $\mathrm{X}$ and testing in $\mathrm{Y}$, while ``w/'' and ``w/o'' denote execution with and without ComManip. Demonstration budgets vary by policy–task pair as specified in Table~\ref{tab:task_training}; within each comparison, the baseline and ComManip use matched budgets.

\begin{wrapfigure}[14]{r}{0.53\textwidth}
    \centering
    \vspace{-10pt}
    \includegraphics[width=\linewidth]{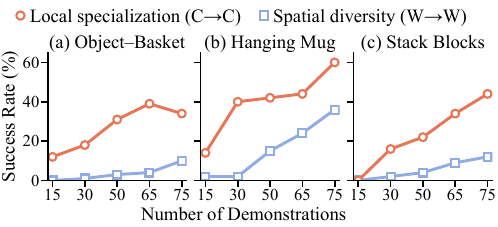}
    \vspace{-8pt}
    \caption{ACT success rates under matched demonstration budgets for comfortable-region and spatially diverse training and evaluation.}
    \label{figdataquantity}
    \vspace{-8pt}
\end{wrapfigure}

\subsection{Data Efficiency under Spatial Diversity}
\label{sec:exp-motivate}
Training within comfortable regions is more data-efficient than training across spatially diverse target placements. We compare policies trained and evaluated within comfortable regions and across spatially diverse target placements, respectively. Under matched demonstration budgets, comfortable-region training (orange in Figure~\ref{figdataquantity}) consistently achieves higher success rates than spatially diverse training (blue) across all three tasks. Notably, on Object Basket, 15 local demonstrations achieve performance comparable to spatially diverse training using 5$\times$ as many demonstrations.

\begin{table}[t]
\centering
\small
\setlength{\tabcolsep}{3.6pt}
\caption{Task success rates (\%) under spatially diverse target placements across the tabletop for six manipulation tasks and five policy families. $\Delta_{\mathrm{avg}}$ denotes the average improvement.}
\label{tabmainresults}

\begin{tabular}{lcccccccl}
\toprule
Policy &
w ComManip &
\shortstack{Stamp\\Seal} &
\shortstack{Empty\\Cup} &
\shortstack{Object\\Basket} &
\shortstack{Phone\\Stand} &
\shortstack{Hanging\\Mug} &
\shortstack{Stack\\Blocks} &
\shortstack{Avg. ($\Delta_{\mathrm{avg}}$)} \\
\midrule

ACT
& -- & 1.6 & 17.4 & 5.6 & 19.0 & 14.0 & 8.6 & 11.0 \\
ACT
& \checkmark
& \textbf{69.6}
& \textbf{82.4}
& \textbf{36.0}
& \textbf{62.8}
& \textbf{62.8}
& \textbf{41.6}
& \textbf{59.2} (+48.2) \\
\midrule

$\pi$0.5
& -- & 48.4 & 57.8 & 9.6 & 42.2 & 16.8 & 38.0 & 35.5 \\
$\pi$0.5
& \checkmark
& \textbf{65.2}
& \textbf{94.4}
& \textbf{24.4}
& \textbf{54.8}
& \textbf{30.8}
& \textbf{58.4}
& \textbf{54.7} (+19.2) \\
\midrule

RDT
& -- & 6.8 & 29.0 & 5.8 & 16.8 & 19.8 & 0.6 & 13.1 \\
RDT
& \checkmark
& \textbf{75.2}
& \textbf{95.2}
& \textbf{28.0}
& \textbf{63.2}
& \textbf{44.4}
& \textbf{48.0}
& \textbf{59.0} (+45.9) \\
\midrule

OpenVLA-OFT
& -- & 16.8 & 6.8 & 2.2 & 10.2 & 27.2 & 1.6 & 10.8 \\
OpenVLA-OFT
& \checkmark
& \textbf{49.2}
& \textbf{97.2}
& \textbf{27.2}
& \textbf{80.4}
& \textbf{43.2}
& \textbf{22.4}
& \textbf{53.3} (+42.5) \\
\midrule

SmolVLA
& -- & 0.8 & 2.2 & 2.0 & 10.8 & 6.2 & 1.6 & 3.9 \\
SmolVLA
& \checkmark
& \textbf{55.6}
& \textbf{65.6}
& \textbf{44.0}
& \textbf{79.2}
& \textbf{31.6}
& \textbf{40.4}
& \textbf{52.7} (+48.8) \\
\midrule

\textbf{Avg. Gain (pp)}
& --
& +48.08
& +64.32
& +26.88
& +48.28
& +25.76
& +32.08
& +40.90 \\
\bottomrule

\end{tabular}
\vspace{-4mm}
\end{table}

\subsection{Main Result}
\label{sec:exp-policies}
ComManip consistently improves performance across different manipulation tasks and policy architectures under spatially diverse target placements. Table~\ref{tabmainresults} compares ComManip with policies trained directly on spatially diverse target placements under matched demonstration budgets. Across five policy families, ComManip consistently improves performance, with an average gain of 40.9 percentage points (pp). Despite achieving the highest baseline performance under spatially diverse target placements, $\pi_{0.5}$ gains a further 19.2 pp with ComManip. The gains span action chunking, diffusion, flow matching, and vision-language-action policies, demonstrating that ComManip is effective across different policy families.

ComManip also improves performance across all tasks, with gains varying in magnitude. The final row of Table~\ref{tabmainresults} shows that the gains averaged across policies range from 25.8 to 64.3 pp. These gains indicate that ComManip is effective across tasks with different manipulation requirements. Overall, these results support the central design of ComManip, which specializes a policy within a comfortable manipulation region and extends it to these placements through test-time repositioning. The consistent gains across tasks and policies demonstrate the broad effectiveness of this strategy under limited demonstration budgets.

\subsection{Effect of Repositioning}
\label{sec:exp-alignment}
\begin{figure*}[t]
    \centering

    \begin{minipage}[t]{0.46\textwidth}
        \vspace{0pt}
        \centering
        \includegraphics[width=\linewidth]
        {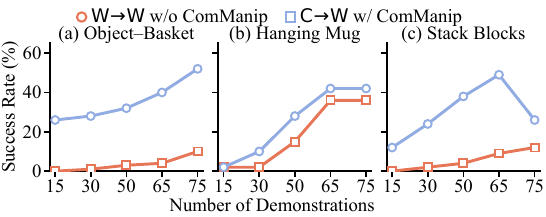}
        \caption{Effect of demonstration quantity on ACT performance under matched demonstration budgets.}
        \label{figdemobudget}
    \end{minipage}
    \hfill
    \begin{minipage}[t]{0.51\textwidth}
        \vspace{0pt}
        \centering
        \captionof{table}{ACT success rates (\%) under different training, evaluation, and deployment settings.}
        \label{tab:act-comparison}
        \vspace{2mm}

        \resizebox{\linewidth}{!}{
        \begin{tabular}{lcccc}
            \toprule
            Task
            & C$\rightarrow$C w/o
            & C$\rightarrow$C w/
            & W$\rightarrow$W w/o
            & C$\rightarrow$W w/ \\
            \midrule
            Stamp Seal  & 88.0 & 98.0 & 1.6  & 69.6 \\
            Empty Cup   & 91.6 & 86.0 & 17.4 & 82.4 \\
            Phone Stand & 52.0 & 76.0 & 19.0 & 62.8 \\
            \midrule
            Average     & 77.2 & 86.7 & 12.7 & 71.6 \\
            \bottomrule
        \end{tabular}
        }
    \end{minipage}
\end{figure*}

\paragraph{Performance within the Comfortable Region.}
Using the same checkpoint, ComManip increases average success from 77.2\% to 86.7\% across three tasks (Table~\ref{tab:act-comparison}). Although Empty Cup shows a modest decrease, its performance remains close to direct execution. These results indicate that repositioning preserves overall local policy performance.

\paragraph{Generalization to Spatially Diverse Placements.}
ComManip achieves 71.6\% average success across spatially diverse target placements, compared with 12.7\% for direct training on these placements (Table~\ref{tab:act-comparison}). This substantial improvement demonstrates that repositioning enables locally trained policies to operate effectively across a broader workspace.

\begin{figure*}[t]
    \centering

    \begin{minipage}[t]{0.46\textwidth}
        \vspace{0pt}
        \centering
        \includegraphics[width=\linewidth]{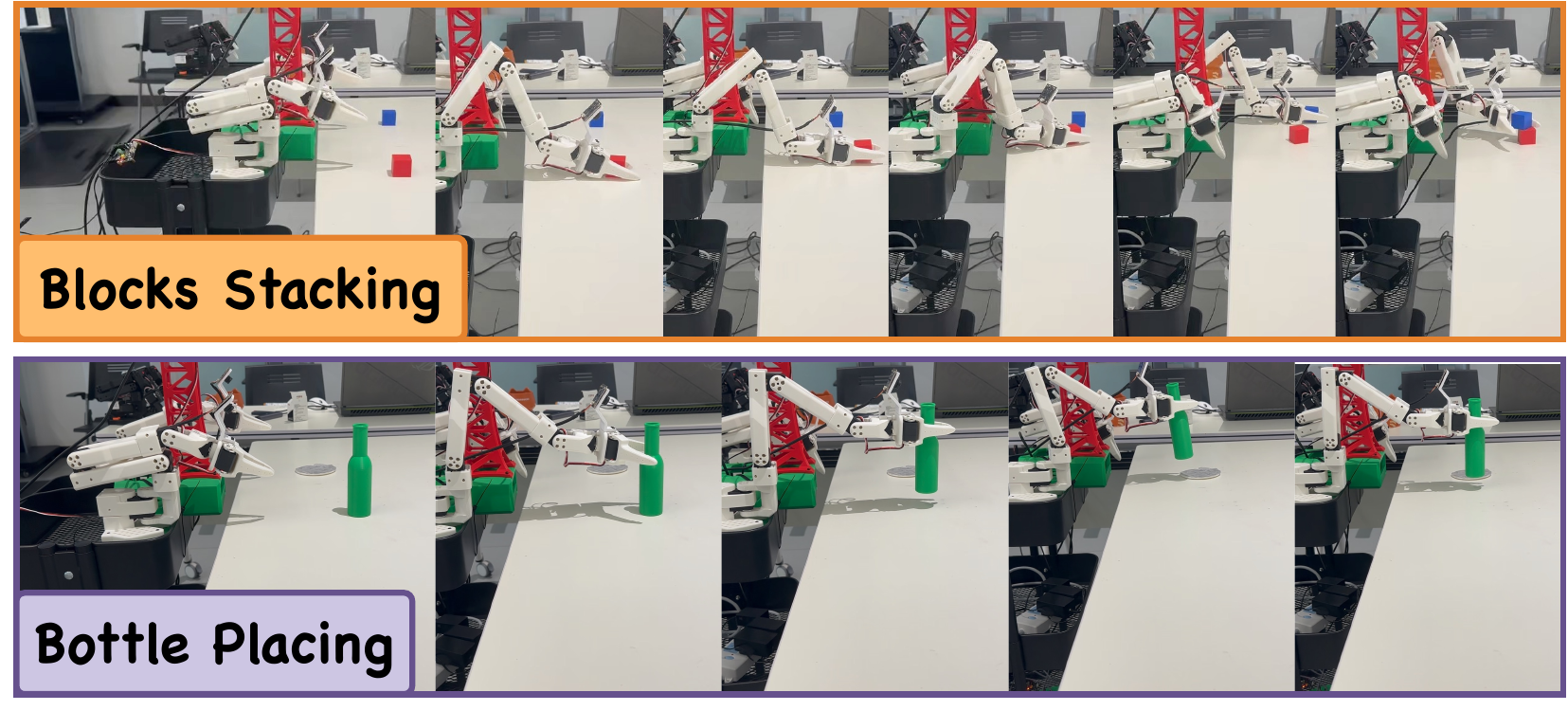}
        \caption{Real-world evaluation tasks using the mobile manipulation platform.}
        \label{fig:real-world}
    \end{minipage}
    \hfill
    \begin{minipage}[t]{0.51\textwidth}
        \vspace{0pt}
        \centering
        \captionof{table}{ACT success rates (\%) in real-world experiments with and without ComManip.}
        \label{tab:real-world}
        \vspace{2mm}
        \small
        \resizebox{\linewidth}{!}{\begin{tabular}{lcc}
            \toprule
            Task & w/o ComManip & w/ ComManip \\
            \midrule
            Bottle Placing & 10.0 & 73.3 \\
            Blocks Stacking  & 6.7 & 56.7 \\
            \midrule
            Average         & 8.3 & 65.0 \\
            \bottomrule
        \end{tabular}}
    \end{minipage}
\end{figure*}

\paragraph{Real-World Validation.}
Real-world experiments with ACT further validate ComManip: success rates increase from 10.0\% to 73.3\% on Bottle Placing and from 6.7\% to 56.7\% on Blocks Stacking (Figure~\ref{fig:real-world} and Table~\ref{tab:real-world}).

\subsection{Effect of Comfortable-Region Extent}
\label{sec:exp-spatial}
We examine the effect of comfortable-region extent under a fixed demonstration budget using two complementary expansion settings.

\begin{figure}[t]
    \centering
    \includegraphics[width=\linewidth]{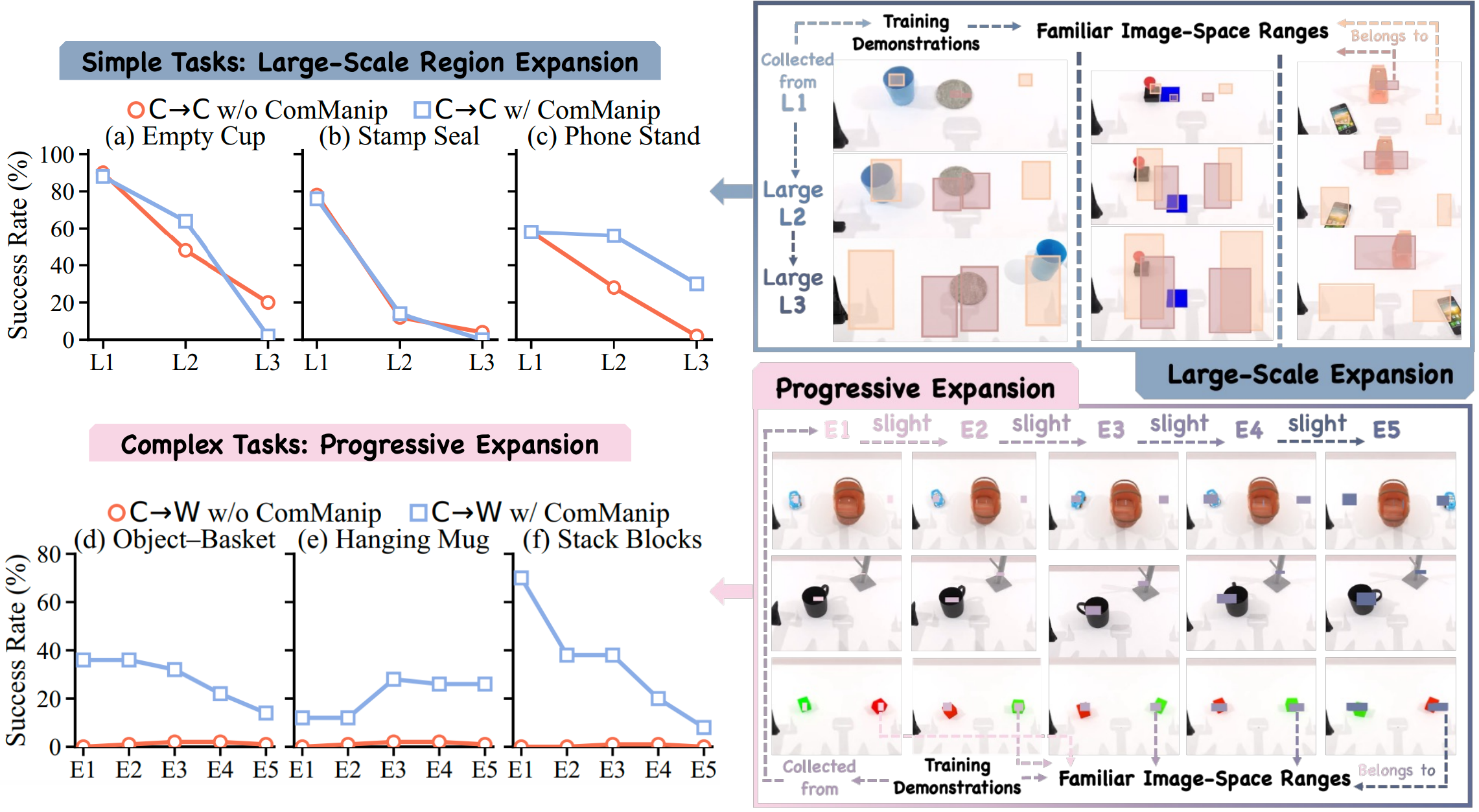}
    \vspace{-4mm}
    \caption{\textbf{Effect of comfortable-region extent on ACT performance.}
    The top row evaluates execution within the comfortable region under
    large-scale expansion, while the bottom row evaluates execution from
    the comfortable region to the full workspace under progressive expansion.
    The corresponding training demonstrations and familiar image-space
    ranges are shown on the right.}
    \label{figcomfortableextentcurve}
    \vspace{-2mm}
\end{figure}

\paragraph{Large-Scale Expansion.}
The first setting evaluates simple tasks as the region expands from L1 to L3 toward the workspace boundary. Performance generally decreases with increasing region extent (Figure~\ref{figcomfortableextentcurve}, top row). ComManip remains comparable to or better than direct execution at L2, whereas both settings degrade at L3 across all tasks. This indicates that excessive expansion weakens local policy learning and limits the benefits of ComManip.

\paragraph{Local Expansion.}
The second setting evaluates progressive local expansions from E1 to E5 on complex tasks. ComManip consistently outperforms direct execution across all region extents (Figure~\ref{figcomfortableextentcurve}, bottom row). Hanging Mug peaks at E3 rather than E1, suggesting that moderate spatial variation can benefit some tasks. These results indicate that effective comfortable regions should balance demonstration concentration with sufficient local variation.

\subsection{Effect of Demonstration Quantity}
\label{sec:exp-data}
ComManip remains effective across a wide range of demonstration budgets, indicating that its advantage is not limited to a particular budget. Its performance generally improves as the number of comfortable-region demonstrations increases. At higher budgets, the trend becomes task-dependent. 
			
As shown in Figure~\ref{figdemobudget}, ComManip matches or exceeds spatially diverse training across all evaluated tasks and demonstration budgets. Its performance generally improves as the number of comfortable-region demonstrations increases. In particular, on Object Basket, increasing the demonstration budget from 15 to 75 raises the success rate to 2$\times$ its value at the initial budget. The effect of additional demonstrations differs across tasks. Some tasks continue to improve, while others show limited improvement or decline at higher budgets. The consistent advantage over spatially diverse training shows that ComManip remains data-efficient across budgets.

\section{Conclusion}
This paper shows that, under limited demonstration budgets, policies trained on spatially diverse target placements are less data-efficient than locally specialized policies trained within compact and stable regions. Motivated by this finding, we introduce ComManip, which combines comfortable-region policy learning with base repositioning to recover familiar visual and kinematic conditions during deployment. Experiments across six tasks and five policy families demonstrate consistent improvements under matched demonstration budgets, while further analyses reveal the effects of repositioning, comfortable-region extent, and demonstration quantity. These results support combining local policy learning with base repositioning to handle target-location variation.

\subsection*{AI use statement}

Generative AI tools were used to assist with manuscript editing, LaTeX reformatting, reference organization, and figure layout planning. The authors reviewed all AI-assisted work and take responsibility for the final text, claims, citations, and artifacts.

\subsection*{Reproducibility statement}

We provide an overview of the ComManip framework and experimental setup in the main paper. The appendix provides further details on comfortable-region construction, familiar image-space estimation, policy training, task configurations, deployment settings, and evaluation protocols to support reproducibility.

\bibliographystyle{arxiv_preprint}
\bibliography{commanip}

\appendix
\section{Experimental Setup}
\subsection{Simulation Environment}
We conduct experiments in RoboTwin 2.0 using the fixed-base
AgileX ALOHA embodiment and the SAPIEN physics backend.
The robot consists of two 6-DoF manipulators, each equipped
with a parallel gripper. We use RoboTwin's default D435
head-camera configuration, which provides RGB observations
at a resolution of $320\times240$ pixels with a vertical
field of view of $37^\circ$. All physical distances are reported in
meters (m). The camera is initialized at
$(-0.032,-0.45,1.35)\,\mathrm{m}$, with forward and left
direction vectors $(0,0.6,-0.8)$ and $(-1,0,0)$,
respectively. Its near and far clipping planes are
$0.1\,\mathrm{m}$ and $100\,\mathrm{m}$. Unless otherwise
stated, the camera orientation, robot embodiment, object
models, physical parameters, and rendering configuration
follow the official RoboTwin 2.0 defaults.

\subsection{Base-Repositioning Implementation}
The AgileX ALOHA embodiment in RoboTwin is fixed-base and
does not provide mobile-base dynamics. We emulate base repositioning by kinematically translating the robot and camera assembly within the tabletop plane.
The head camera and associated light root are translated by
the same displacement while preserving their relative
transforms and camera orientation.

Each movement uses a task- and stage-specific step size
$s\in\{0.002,0.005\}$~m and is interpolated over eight
simulation frames. When initial object normalization is
enabled, two additional settling steps are executed. 

\subsection{Evaluation Protocol}

Evaluation follows the official RoboTwin protocol. Unless
otherwise specified, success is determined by the
task-specific \texttt{check\_success()} implementation.

For Stamp Seal, we use a planar position tolerance of
$\epsilon_{\mathrm{stamp}}=0.02\,\mathrm{m}$.
Here,
$(x_{\mathrm{stamp}},y_{\mathrm{stamp}})$ and
$(x_{\mathrm{target}},y_{\mathrm{target}})$ denote the
planar physical positions of the stamp and target region,
respectively, in the RoboTwin world coordinate frame.
An episode is considered successful when both grippers are
open and the following planar conditions hold:
\[
\left|x_{\mathrm{stamp}}-x_{\mathrm{target}}\right|
<\epsilon_{\mathrm{stamp}},
\qquad
\left|y_{\mathrm{stamp}}-y_{\mathrm{target}}\right|
<\epsilon_{\mathrm{stamp}}.
\]
This criterion is applied consistently to all policies,
baselines, and ComManip variants.

If the success condition is not satisfied, the episode terminates when the task-specific maximum action-step limit is reached. Success is evaluated automatically without manual intervention. The official maximum action-step limits for the evaluated tasks are
summarized in Table~\ref{tab:evaluation-steps}.

\begin{table}[t]
\centering
\small
\caption{Official RoboTwin maximum action-step limits for the evaluated tasks.}
\label{tab:evaluation-steps}
\begin{tabular}{llc}
\toprule
Task & RoboTwin Task Name & Max. Steps \\
\midrule
Stamp Seal    & \texttt{stamp\_seal}          & 400 \\
Empty Cup     & \texttt{place\_empty\_cup}    & 500 \\
Object Basket & \texttt{place\_object\_basket} & 700 \\
Phone Stand   & \texttt{place\_phone\_stand}  & 400 \\
Hanging Mug   & \texttt{hanging\_mug}         & 900 \\
Stack Blocks  & \texttt{stack\_blocks\_two}   & 800 \\
\bottomrule
\end{tabular}
\end{table}

\subsection{Manipulation Tasks}

The six evaluated RoboTwin tasks and their primary
manipulation challenges are summarized in
Table~\ref{tab:task_details}.

\begin{table*}[t]
\centering
\small
\setlength{\tabcolsep}{4pt}
\renewcommand{\arraystretch}{1.5}
\caption{Evaluated manipulation tasks and their primary
technical challenges.}
\label{tab:task_details}
\begin{tabularx}{\textwidth}{
    l
    >{\raggedright\arraybackslash}p{4cm}
    >{\raggedright\arraybackslash}p{2.2cm}
    >{\raggedright\arraybackslash}X}
\toprule
\textbf{Task} &
\textbf{Skill Type} &
\textbf{Object / Target} &
\textbf{Main Challenge} \\
\midrule

Stamp Seal &
Planar alignment and contact-rich pressing &
Stamp and target region &
Accurately aligning the stamp with a small target region
before establishing downward contact. \\

Empty Cup &
Grasp-and-place with placement-sensitive release &
Cup and coaster &
Placing the cup stably on a spatially restricted support region without excessive positional or height error. \\

Object Basket &
Bimanual grasp-and-place &
Object and basket &
Coordinating object transfer and basket manipulation while keeping the basket upright and avoiding contact with the table. \\

Phone Stand &
Orientation-sensitive constrained placement &
Phone and stand &
Aligning the phone with the narrow supporting region of the stand before release. \\

Hanging Mug &
Grasping and hook alignment &
Mug and rack &
Aligning the mug handle with the rack while maintaining sufficient height for successful hanging. \\

Stack Blocks &
Sequential stacking and precision placement &
Multiple blocks &
Preventing positional errors from accumulating across sequential grasping and stacking stages. \\

\bottomrule
\end{tabularx}
\end{table*}

\subsection{Policies}
For ACT, $\pi_{0.5}$, RDT, and OpenVLA-OFT, we use the
implementations provided by RoboTwin 2.0. SmolVLA is
integrated from its released base checkpoint using the
recommended fine-tuning configuration.

All policies receive three RGB streams from one head camera
and two wrist cameras. Actions are executed through the same
14-dimensional RoboTwin joint-position interface.
Model-specific preprocessing and internal representations are
otherwise retained.

\subsection{Demonstration Collection}

Demonstrations are generated using RoboTwin's automated
expert framework. All tasks inherit from
\texttt{Base\_Task} and implement task-specific
\texttt{play\_once()} procedures by composing a shared set
of low-level manipulation primitives.

Only successful expert rollouts are retained. For each task
and placement distribution, the retained episodes form a
common demonstration pool shared by all policy families.
Policies with matched demonstration budgets use the same
underlying episodes, while policies with different budgets
use fixed subsets of the corresponding pool. Comfortable-
region and workspace-wide variants use matched numbers of
successful demonstrations, as reported in
Table~\ref{tab:task_training}.

\subsection{Target-Placement Distributions}
Candidate object positions are sampled uniformly from the
task-specific bounds listed in
Tables~\ref{tab:comfortable_ranges}
and~\ref{tab:wide_ranges}, and are filtered using the
corresponding RoboTwin validity checks. Object orientations
follow the official task configurations.
\begin{table*}[t]
\centering
\small
\setlength{\tabcolsep}{5pt}
\renewcommand{\arraystretch}{1.2}
\caption{Object-placement ranges for the comfortable-region
distributions. Positions are expressed in the RoboTwin world
coordinate frame. Candidate positions are sampled uniformly
within the listed ranges.}
\label{tab:comfortable_ranges}
\begin{tabular}{
    C{3cm}
    C{3cm}
    C{4.5cm}
    C{4.5cm}}
\toprule
\textbf{Task} &
\textbf{Varied Object} &
\textbf{$x$ Range (m)} &
\textbf{$y$ Range (m)} \\
\midrule

Stamp Seal
& Stamp
& $[-0.15 ,-0.1], [0.1 ,0.15]$
& $[-0.025 ,0.025]$ \\

& Target
& $[-0.09, -0.04],[0.04, 0.09]$
& $[-0.025, 0.025]$\\

Empty Cup
& Cup
& $[-0.2, -0.15], [0.15, 0.2]$
& $[0,0.05]$ \\
& Coaster
& $[-0.02, 0.03], [-0.03, 0.02]$
& $[0,0.05]$ \\

Object Basket
& Object
& $[-0.25, -0.2], [0.2, 0.25]$
& $[-0.025, 0.025]$ \\
& Basket
& $[0.02, 0.02]$
& $[-0.08, -0.05]$ \\

Phone Stand
& Phone
& $[-0.2, -0.15], [0.15, 0.2]$
& $[-0.1, 0.05]$ \\
& Stand
& $[-0.05, 0.0], [0, 0.05]$
& $[0.02, 0.07]$ \\

Hanging Mug
& Mug
& $[-0.15, -0.1]$
& $[-0.03, 0.03]$  \\

& Rack
& $[0.1, 0.15]$
& $[0.13, 0.16]$ \\

Stack Blocks
& Block 1 / Block 2
& $[-0.2, -0.15], [0.15, 0.2]$
& $[-0.04, 0.01]$  \\

\bottomrule
\end{tabular}
\end{table*}

\begin{table*}[t]
\centering
\small
\setlength{\tabcolsep}{5pt}
\renewcommand{\arraystretch}{1.2}
\caption{Object-placement ranges for the workspace-wide
distributions. Positions are expressed in the RoboTwin world
coordinate frame. Candidate positions are sampled uniformly
within the listed ranges.}
\label{tab:wide_ranges}
\begin{tabular}{
    C{3cm}
    C{3cm}
    C{4.5cm}
    C{4.5cm}}
\toprule
\textbf{Task} &
\textbf{Varied Object} &
\textbf{$x$ Range (m)} &
\textbf{$y$ Range (m)} \\
\midrule

\multirow[c]{2}{3cm}{\centering Stamp Seal}
& Stamp
& $[-0.25, -0.1], [0.1, 0.25]$
& $[-0.25, 0.25]$ \\
& Target
& $[-0.19, -0.04], [0.04, 0.19]$
& $[-0.25, 0.25]$ \\

\multirow[c]{2}{3cm}{\centering Empty Cup}
& Cup
& $[-0.3, -0.15], [0.15, 0.3]$
& $[-0.2, 0.05]$\\
& Coaster
& $[-0.05, 0.1], [-0.1, 0.05]$
& $[-0.2, 0.05]$ \\

\multirow[c]{2}{3cm}{\centering Object Basket}
& Object
& $[-0.35, -0.15], [0.15, 0.35]$
& $[-0.25, 0.25]$ \\

& Basket
& $[0.02, 0.02]$
& $[-0.08, -0.02]$ \\

\multirow[c]{2}{3cm}{\centering Phone Stand}
& Phone
& $[-0.25, -0.05], [0.05, 0.25]$
& $[-0.2, 0]$ \\
& Stand
& $[-0.15, 0], [0, 0.15]$
& $[0, 0.2]$ \\

\multirow[c]{2}{3cm}{\centering Hanging Mug}
& Mug
& $[-0.25, -0.1]$
& $[-0.05, 0.05]$ \\
& Rack
& $[0.1, 0.3]$
& $[0.13, 0.17]$ \\

\multirow[c]{1}{3cm}{\centering Stack Blocks}
& Block 1 / Block 2
& $[-0.28,-0.15], [0.15,0.28]$
& $[-0.08, 0.05]$  \\
\bottomrule
\end{tabular}
\end{table*}

\subsection{Comfortable-Region Extent Settings}
Tables~\ref{tab:extent_l123} and \ref{tab:extent_e12345} list the physical target-placement bounds used in the comfortable-region extent study. L1, L2, and L3 are the three large-scale settings for the simple tasks. E1 to E5 are the progressive settings for the complex tasks. For multi-object tasks, all listed object ranges are varied jointly within each setting.

\begin{table}[t]
\centering
\caption{Physical target-placement ranges used for the
large-scale comfortable-region expansion from L1 to L3.
Each cell reports the $x$ and $y$ ranges in the RoboTwin
world coordinate frame.}
\label{tab:extent_l123}

\resizebox{\columnwidth}{!}{%
\begin{tabular}{ccccc}
\toprule
\textbf{Task} &
\textbf{Varied Object} &
\textbf{L1 (m)} &
\textbf{L2 (m)} &
\textbf{L3 (m)} \\
\midrule

\multirow[c]{2}{*}{Empty Cup}
& Cup
& \shortstack{$x:[-0.2,-0.15], [0.15,0.2]$\\$y:[0,0.05]$}
& \shortstack{$x:[-0.25,-0.15], [0.15,0.25]$\\$y:[-0.1,0.05]$}
& \shortstack{$x:[-0.3,-0.15], [0.15,0.3]$\\$y:[-0.2,0.05]$} \\

& Coaster
& \shortstack{$x:[-0.02,0.03], [-0.03,0.02]$\\$y:[0,0.05]$}
& \shortstack{$x:[-0.02,0.08], [-0.08,0.02]$\\$y:[-0.1,0.05]$}
& \shortstack{$x:[-0.05,0.1], [-0.1,0.05]$\\$y:[-0.2,0.05]$} \\

\midrule

\multirow[c]{2}{*}{Stamp Seal}
& Stamp
& \shortstack{$x:[-0.15,-0.1], [0.1,0.15]$\\$y:[-0.025,0.025]$}
& \shortstack{$x:[-0.2,-0.1], [0.1,0.2]$\\$y:[-0.125,0.125]$}
& \shortstack{$x:[-0.25,-0.1], [0.1,0.25]$\\$y:[-0.25,0.25]$} \\

& Target
& \shortstack{$x:[-0.09,-0.04], [0.04,0.09]$\\$y:[-0.025,0.025]$}
& \shortstack{$x:[-0.14,-0.09], [0.09,0.14]$\\$y:[-0.125,0.125]$}
& \shortstack{$x:[-0.19,-0.04], [0.04,0.19]$\\$y:[-0.25,0.25]$} \\

\midrule

\multirow[c]{2}{*}{Phone Stand}
& Phone
& \shortstack{$x:[-0.2,-0.15], [0.15,0.2]$\\$y:[-0.1,0.05]$}
& \shortstack{$x:[-0.25,-0.15], [0.15,0.25]$\\$y:[-0.2,0.0]$}
& \shortstack{$x:[-0.25,-0.05], [0.05,0.25]$\\$y:[-0.2,0.0]$} \\

& Stand
& \shortstack{$x:[-0.05,0], [0,0.05]$\\$y:[0.02,0.07]$}
& \shortstack{$x:[-0.1,0.0], [0,0.1]$\\$y:[0.02,0.12]$}
& \shortstack{$x:[-0.15,0.0], [0,0.15]$\\$y:[0,0.2]$} \\

\bottomrule
\end{tabular}%
}
\end{table}

\begin{table}[t]
\centering
\caption{Physical target-placement ranges used for the
progressive comfortable-region expansion from E1 to E5.
Each cell reports the $x$ and $y$ ranges in the RoboTwin
world coordinate frame. All relevant object ranges are
varied jointly within each setting.}
\label{tab:extent_e12345}

\resizebox{\columnwidth}{!}{%
\begin{tabular}{cc*{5}{c}}
\toprule
\textbf{Task} &
\textbf{Varied Object} &
\textbf{E1 (m)} &
\textbf{E2 (m)} &
\textbf{E3 (m)} &
\textbf{E4 (m)} &
\textbf{E5 (m)} \\
\midrule

\multirow[c]{2}{*}{Object Basket}
& Object
& \shortstack{
$x:[-0.25,-0.23],$\\
$\phantom{x:}[0.23,0.25]$\\
$y:[-0.025,0.025]$
}
& \shortstack{
$x:[-0.25,-0.22],$\\
$\phantom{x:}[0.22,0.25]$\\
$y:[-0.025,0.025]$
}
& \shortstack{
$x:[-0.25,-0.2],$\\
$\phantom{x:}[0.2,0.25]$\\
$y:[-0.025,0.025]$
}
& \shortstack{
$x:[-0.275,-0.22],$\\
$\phantom{x:}[0.22,0.275]$\\
$y:[-0.03,0.03]$
}
& \shortstack{
$x:[-0.275,-0.22],$\\
$\phantom{x:}[0.22,0.275]$\\
$y:[-0.04,0.04]$
} \\

& Basket
& \shortstack{$x:[0.02,0.02]$\\$y:[-0.06,-0.05]$}
& \shortstack{$x:[0.02,0.02]$\\$y:[-0.07,-0.05]$}
& \shortstack{$x:[0.02,0.02]$\\$y:[-0.08,-0.05]$}
& \shortstack{$x:[0.02,0.02]$\\$y:[-0.09,-0.05]$}
& \shortstack{$x:[0.02,0.02]$\\$y:[-0.1,-0.05]$} \\

\midrule

\multirow[c]{2}{*}{Hanging Mug}
& Mug
& \shortstack{$x:[-0.125,-0.1]$\\$y:[-0.01,0.01]$}
& \shortstack{$x:[-0.15,-0.1]$\\$y:[-0.015,0.015]$}
& \shortstack{$x:[-0.15,-0.1]$\\$y:[-0.03,0.03]$}
& \shortstack{$x:[-0.175,-0.1]$\\$y:[-0.03,0.03]$}
& \shortstack{$x:[-0.175,-0.1]$\\$y:[-0.04,0.04]$} \\

& Rack
& \shortstack{$x:[0.1,0.12]$\\$y:[0.13,0.14]$}
& \shortstack{$x:[0.1,0.14]$\\$y:[0.13,0.15]$}
& \shortstack{$x:[0.1,0.15]$\\$y:[0.13,0.16]$}
& \shortstack{$x:[0.1,0.15]$\\$y:[0.13,0.17]$}
& \shortstack{$x:[0.1,0.16]$\\$y:[0.13,0.18]$} \\

\midrule

Stack Blocks
& \makecell[c]{Block 1\\Block 2}
& \shortstack{
$x:[-0.2,-0.18],$\\
$\phantom{x:}[0.18,0.2]$\\
$y:[-0.04,0.01]$
}
& \shortstack{
$x:[-0.2,-0.16],$\\
$\phantom{x:}[0.16,0.2]$\\
$y:[-0.04,0.01]$
}
& \shortstack{
$x:[-0.2,-0.15],$\\
$\phantom{x:}[0.15,0.2]$\\
$y:[-0.04,0.01]$
}
& \shortstack{
$x:[-0.23,-0.15],$\\
$\phantom{x:}[0.15,0.23]$\\
$y:[-0.04,0.01]$
}
& \shortstack{
$x:[-0.25,-0.15],$\\
$\phantom{x:}[0.15,0.25]$\\
$y:[-0.04,0.01]$
} \\

\bottomrule
\end{tabular}%
}
\end{table}

\section{Implementation Details}

\subsection{Object Localization and Segmentation}
\begin{figure*}[t]
\centering
\begingroup
\renewcommand{\tabularxcolumn}[1]{m{#1}}
\setlength{\tabcolsep}{3pt}
\renewcommand{\arraystretch}{1.15}

\begin{tabularx}{\textwidth}{
    >{\centering\arraybackslash}m{1.8cm}
    >{\centering\arraybackslash}X
    >{\centering\arraybackslash}X
    >{\centering\arraybackslash}X}
\toprule

\textbf{Task} &
\textbf{GroundingDINO Detections and Scores} &
\textbf{SAM 2.1 Mask for the Pick Target} &
\textbf{SAM 2.1 Mask for the Place Target} \\

\midrule

Stamp Seal
&
\includegraphics[
    width=\linewidth,
    height=2.15cm,
    keepaspectratio
]{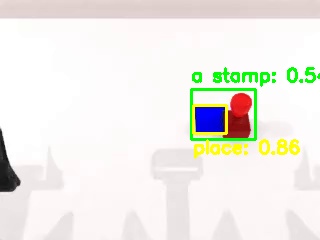}
&
\includegraphics[
    width=\linewidth,
    height=2.15cm,
    keepaspectratio
]{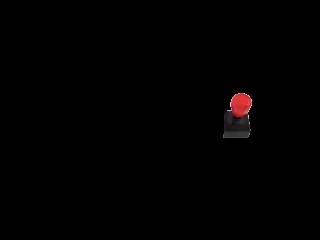}
&
\includegraphics[
    width=\linewidth,
    height=2.15cm,
    keepaspectratio
]{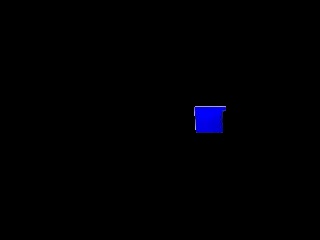}
\\

\midrule

Empty Cup
&
\includegraphics[
    width=\linewidth,
    height=2.15cm,
    keepaspectratio
]{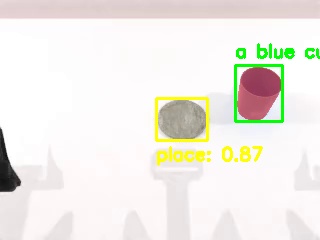}
&
\includegraphics[
    width=\linewidth,
    height=2.15cm,
    keepaspectratio
]{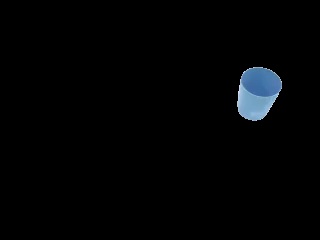}
&
\includegraphics[
    width=\linewidth,
    height=2.15cm,
    keepaspectratio
]{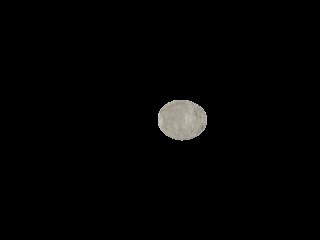}
\\

\midrule

Object Basket
&
\includegraphics[
    width=\linewidth,
    height=2.15cm,
    keepaspectratio
]{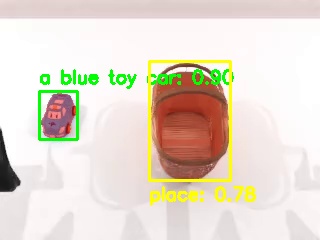}
&
\includegraphics[
    width=\linewidth,
    height=2.15cm,
    keepaspectratio
]{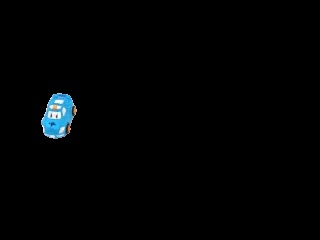}
&
\includegraphics[
    width=\linewidth,
    height=2.15cm,
    keepaspectratio
]{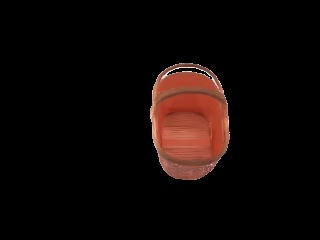}
\\

\midrule

Phone Stand
&
\includegraphics[
    width=\linewidth,
    height=2.15cm,
    keepaspectratio
]{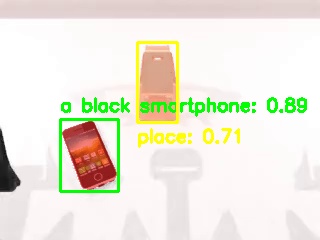}
&
\includegraphics[
    width=\linewidth,
    height=2.15cm,
    keepaspectratio
]{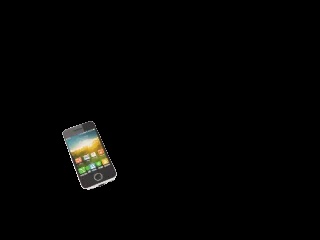}
&
\includegraphics[
    width=\linewidth,
    height=2.15cm,
    keepaspectratio
]{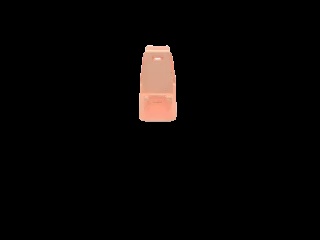}
\\

\midrule

Hanging Mug
&
\includegraphics[
    width=\linewidth,
    height=2.15cm,
    keepaspectratio
]{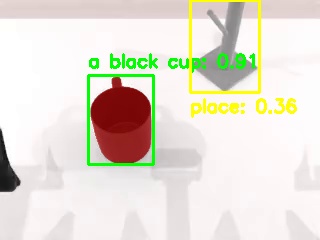}
&
\includegraphics[
    width=\linewidth,
    height=2.15cm,
    keepaspectratio
]{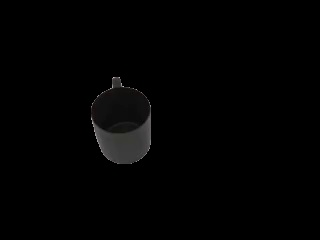}
&
\includegraphics[
    width=\linewidth,
    height=2.15cm,
    keepaspectratio
]{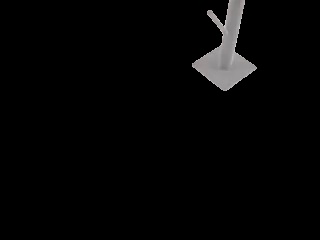}
\\

\midrule

\multirow{2}{1.8cm}{%
    \parbox[c][2.5cm][c]{1.8cm}{%
        \centering
        Stack\\Blocks
    }%
}
&
\includegraphics[
    width=\linewidth,
    height=2.15cm,
    keepaspectratio
]{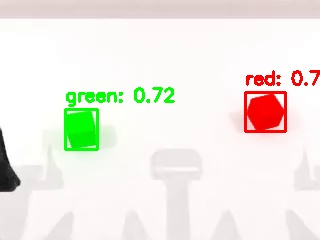}
&
\includegraphics[
    width=\linewidth,
    height=2.15cm,
    keepaspectratio
]{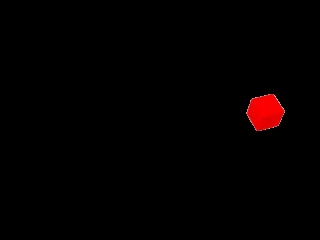}
&
\textemdash
\\

&
\includegraphics[
    width=\linewidth,
    height=2.15cm,
    keepaspectratio
]{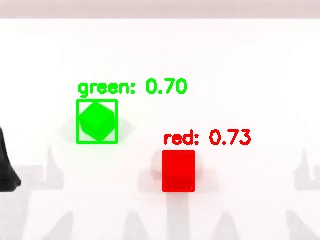}
&
\includegraphics[
    width=\linewidth,
    height=2.15cm,
    keepaspectratio
]{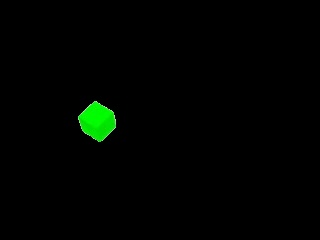}
&
\includegraphics[
    width=\linewidth,
    height=2.15cm,
    keepaspectratio
]{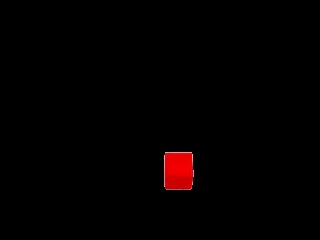}
\\

\bottomrule

\end{tabularx}
\endgroup

\caption{Qualitative localization results across the six evaluated tasks. From left to right, the columns show GroundingDINO detections and confidence scores, pick-target masks, and place-target masks produced by SAM~2.1. The mask centroid is used as the image-space target center. For Stack Blocks, the dash indicates that no separate place-target mask is required during the first pick stage.}
\label{fig:localization_examples}
\end{figure*}

Given the current head-camera RGB observation and a
task- and stage-specific target prompt,
GroundingDINO-SwinT predicts candidate bounding boxes
and confidence scores. Detections are retained when both their box- and text-confidence scores exceed 0.25. For the target relevant to the current stage, the highest-confidence retained box is selected. The selected box is passed to SAM 2.1 Hiera-Small as a box prompt, and the centroid of the resulting mask is used as the image-space target center
$\mathbf{c}_{\mathrm{test}}
=(u_{\mathrm{test}},v_{\mathrm{test}})$,
where $u_{\mathrm{test}}$ and $v_{\mathrm{test}}$ denote the
horizontal and vertical pixel coordinates, respectively.
Qualitative detection and segmentation results for all six
tasks are shown in Figure~\ref{fig:localization_examples}.

GroundingDINO provides semantic target selection, whereas
SAM~2.1 refines the selected box for mask-based center
estimation. This avoids directly relying on a bounding-box
center that may include background or poorly represent an
asymmetric object.

\subsection{Image-to-World Movement Mapping}
The familiar-range construction and closed-loop alignment procedure follow the main paper. In the RoboTwin world frame, image-space corrections are converted to planar robot translations as
\begin{equation}
\begin{aligned}
\Delta p_x &=
\begin{cases}
-s, & u_{\mathrm{test}} < u_{\min},\\
 s, & u_{\mathrm{test}} > u_{\max},\\
 0, & \text{otherwise},
\end{cases}
\\[3pt]
\Delta p_y &=
\begin{cases}
 s, & v_{\mathrm{test}} < v_{\min},\\
-s, & v_{\mathrm{test}} > v_{\max},\\
 0, & \text{otherwise}.
\end{cases}
\end{aligned}
\label{eq:base_movement}
\end{equation}
Here, $\Delta p_x$ and $\Delta p_y$ denote the planar robot translations in the RoboTwin world frame, and $s$ is the task- and stage-specific movement step, selected from
$\{0.002,0.005\}$~m. The quantities $u_{\min}$, $u_{\max}$, $v_{\min}$, and
$v_{\max}$ are the pixel-coordinate bounds of the corresponding familiar image-space range.

When both coordinates lie outside the familiar range, both translation components are applied in the same iteration. After each movement, a new observation is acquired and the target center is re-estimated.

\subsection{Task-Specific Familiar Ranges}
Table~\ref{tab:familiar_ranges} reports the familiar image-space ranges $\mathcal{F}_C$ used during deployment. All coordinates are measured in pixels in the original $320\times240$ head-camera image. Separate ranges are used for different manipulation stages and left/right configurations when applicable.

\begin{table*}[t]
\centering
\small
\setlength{\tabcolsep}{5pt}
\renewcommand{\arraystretch}{2}
\caption{Task- and stage-specific familiar image-space ranges
$\mathcal{F}_C=[u_{\min},u_{\max}]
\times[v_{\min},v_{\max}]$.
L and R denote the left- and
right-side configurations, respectively, while Both indicates
that the same range is used for both configurations.}
\label{tab:familiar_ranges}
\begin{tabular}{
    p{2.2cm}
    p{2.8cm}
    p{1.2cm}
    p{2.5cm}
    p{2.5cm}}
\toprule
\textbf{Task} &
\textbf{Stage / Target} &
\textbf{Side} &
\textbf{$u$ Range (pixels)} &
\textbf{$v$ Range (pixels)} \\
\midrule

\multirow[c]{4}{*}{Empty Cup}
& Cup Pick
& L
& $[75.5,98.0]$
& $[89.0,106.5]$ \\

& Cup Pick
& R
& $[253.5,273.0]$
& $[89.0,107.0]$ \\

& Coaster Place
& L
& $[160.5,175.0]$
& $[112.5,122.0]$ \\

& Coaster Place
& R
& $[176.0,188.5]$
& $[116.0,122.0]$ \\

\midrule

\multirow[c]{3}{*}{Phone Stand}
& Phone Pick
& L
& $[73.0,101.0]$
& $[143.5,161.5]$ \\

& Phone Pick
& R
& $[250.5,281.5]$
& $[142.0,163.0]$ \\

& Stand Place
& Both
& $[151.5,198.0]$
& $[76.5,95.0]$ \\

\midrule

\multirow[c]{4}{*}{Object Basket}
& Object Pick
& L
& $[54.5,78.0]$
& $[106.5,122.5]$ \\

& Object Pick
& R
& $[271.5,294.5]$
& $[106.5,125.5]$ \\

& Basket Place
& L
& $[166.5,167.0]$
& $[117.5,130.0]$ \\

& Basket Place
& R
& $[189.0,190.0]$
& $[118.0,129.0]$ \\

\midrule

\multirow[c]{2}{*}{Hanging Mug}
& Mug Pick
& Both
& $[91.0,125.0]$
& $[105.0,122.0]$ \\

& Rack Place
& Both
& $[220.0,241.5]$
& $[44.0,49.0]$ \\

\midrule

\multirow[c]{5}{*}{Stack Blocks}
& Red Block Pick
& L
& $[78.5,94.5]$
& $[112.0,130.5]$ \\

& Red Block Pick
& R
& $[248.5,274.5]$
& $[112.0,130.5]$ \\

& Green Block Pick
& L
& $[75.5,99.5]$
& $[112.0,130.5]$ \\

& Green Block Pick
& R
& $[249.5,272.0]$
& $[111.5,130.0]$ \\

& Block Place
& Both
& $[176.4,178.2]$
& $[171.7,172.4]$ \\

\midrule

\multirow[c]{4}{*}{Stamp Seal}
& Stamp Pick
& L
& $[58.0,128.5]$
& $[52.0,202.0]$ \\

& Stamp Pick
& R
& $[227.0,302.5]$
& $[46.0,222.5]$ \\

& Target Place
& L
& $[86.5,147.0]$
& $[91.0,219.0]$ \\

& Target Place
& R
& $[207.0,282.0]$
& $[111.5,226.5]$ \\

\bottomrule
\end{tabular}
\end{table*}

\subsection{Failure Handling}
After at least one successful detection, the visual servo tolerates at most three consecutive detection misses. Alignment terminates with \texttt{detection\_failed} after three consecutive detection misses and with \texttt{max\_iters} when the target does not enter $\mathcal{F}_C$ within $K_{\max}$ movements.

\subsection{Policy-State Reset}
After repositioning, we reset each policy before execution to prevent alignment observations or action history from carrying over. This clears the temporal-aggregation state for
ACT, the policy state and action queue for SmolVLA, and the observation window for RDT and $\pi_{0.5}$. OpenVLA-OFT maintains no persistent state in the evaluation wrapper and therefore requires no reset. Each policy then receives a fresh observation from the repositioned camera.

\section{Additional Training Details}
\label{app:training-details}
We retain the original initialization and training procedure of each policy family. ACT is trained from scratch except that its ResNet-18 visual backbone is initialized with ImageNet-pretrained weights. The remaining policies are fine-tuned from their released base checkpoints. The task-specific demonstration budgets, learning rates, global batch sizes, training lengths, gradient-accumulation steps, and evaluated checkpoints are summarized in Table~\ref{tab:task_training}.

\begin{table*}[t]
\centering
\small
\setlength{\tabcolsep}{3.2pt}
\renewcommand{\arraystretch}{1.05}
\caption{Task-specific demonstration budgets and training configurations.
The final training checkpoint is used for evaluation.
For $\pi_{0.5}$, the listed learning rate is the peak learning rate of the training schedule.}
\label{tab:task_training}

\begin{tabular}{llccccc}
\toprule
\textbf{Policy} &
\textbf{Task} &
\textbf{\# Demos} &
\textbf{Learning Rate} &
\textbf{Global Batch} &
\textbf{Training Steps} &
\textbf{Grad. Accum.} \\
\midrule

ACT & All six tasks & 50 per task & $1\times10^{-5}$ & 8 & 5000 & 1 \\
\midrule

\multirow{6}{*}{$\pi_{0.5}$}
& Stamp Seal    & 30 & $2.5\times10^{-5}$ & 4  & 5000 & 1 \\
& Empty Cup     & 30 & $2.5\times10^{-5}$ & 8  & 5000 & 1 \\
& Object Basket & 75 & $2.5\times10^{-5}$ & 16 & 5000 & 1 \\
& Phone Stand   & 30 & $2.5\times10^{-5}$ & 8  & 5000 & 1 \\
& Hanging Mug   & 75 & $2.5\times10^{-5}$ & 16 & 5000 & 1 \\
& Stack Blocks  & 50 & $2.5\times10^{-5}$ & 16 & 5000 & 1 \\
\midrule

\multirow{6}{*}{RDT}
& Stamp Seal    & 30 & $5\times10^{-5}$ & 4  & 20000 & 1 \\
& Empty Cup     & 30 & $5\times10^{-5}$ & 8  & 20000 & 1 \\
& Object Basket & 75 & $5\times10^{-5}$ & 16 & 20000 & 1 \\
& Phone Stand   & 30 & $5\times10^{-5}$ & 8  & 20000 & 1 \\
& Hanging Mug   & 75 & $5\times10^{-5}$ & 16 & 20000 & 1 \\
& Stack Blocks  & 50 & $5\times10^{-5}$ & 16 & 20000 & 1 \\
\midrule

\multirow{6}{*}{OpenVLA-OFT}
& Stamp Seal    & 50 & $1\times10^{-4}$ & 8  & 20000 & 2 \\
& Empty Cup     & 50 & $5\times10^{-5}$ & 8  & 20000 & 2 \\
& Object Basket & 75 & $5\times10^{-5}$ & 16 & 20000 & 4 \\
& Phone Stand   & 50 & $1\times10^{-4}$ & 4  & 20000 & 1 \\
& Hanging Mug   & 75 & $1\times10^{-4}$ & 8  & 20000 & 2 \\
& Stack Blocks  & 75 & $5\times10^{-5}$ & 16 & 20000 & 4 \\
\midrule

\multirow{6}{*}{SmolVLA}
& Stamp Seal    & 50 & $5\times10^{-5}$ & 8  & 20000 & 1 \\
& Empty Cup     & 50 & $5\times10^{-5}$ & 4  & 20000 & 1 \\
& Object Basket & 75 & $5\times10^{-5}$ & 16 & 20000 & 1 \\
& Phone Stand   & 50 & $5\times10^{-5}$ & 4  & 20000 & 1 \\
& Hanging Mug   & 75 & $5\times10^{-5}$ & 8  & 20000 & 1 \\
& Stack Blocks  & 50 & $5\times10^{-5}$ & 8  & 20000 & 1 \\

\bottomrule
\end{tabular}
\end{table*}

\subsection{Computational Resources}
Training is conducted on one NVIDIA H100 GPU with 80~GB of memory. Policy evaluation, including GroundingDINO and SAM~2.1 inference, is performed on one NVIDIA GeForce
RTX~3090 GPU. The batch sizes reported in Table~\ref{tab:task_training} are global batch sizes after gradient accumulation.

\end{document}